\pdfoutput=1

\documentclass[11pt]{article}
\usepackage{authblk}

\usepackage[dvipsnames]{xcolor}

\usepackage[]{acl}

\usepackage{times}
\usepackage{latexsym}

\usepackage[T1]{fontenc}

\usepackage[utf8]{inputenc}

\usepackage{booktabs}
\usepackage{tabularx}
\usepackage{lipsum}
\usepackage{graphicx}
\usepackage{todonotes}
\usepackage{svg}
\usepackage{amsmath}
\usepackage{multirow}
\usepackage{csquotes}
\usepackage{supertabular} 
\usepackage{longtable}
\usepackage{float}

\usepackage{microtype}

\title{Textual Entailment for Event Argument Extraction: \\ Zero- and Few-Shot with Multi-Source Learning}

\author[1]{\textbf{Oscar Sainz}}
\author[1]{\textbf{Itziar Gonzalez-Dios}}
\author[1]{\\\textbf{Oier Lopez de Lacalle}}
\author[2]{\textbf{Bonan Min}}
\author[1]{\textbf{Eneko Agirre}}

\affil[1]{HiTZ Basque Center for Language Technologies - Ixa NLP Group}
\affil[ ]{University of the Basque Country UPV/EHU}
\affil[2]{Raytheon BBN Technologies}
\affil[ ]{ oscar.sainz@ehu.eus}

\date{}

\begin{document}
\maketitle

\begin{abstract}
Recent work has shown that NLP tasks such as 
Relation Extraction (RE) can be recasted as  Textual Entailment tasks using verbalizations, with strong performance in zero-shot and few-shot settings thanks to pre-trained entailment models. The fact that relations in current RE datasets are easily verbalized casts doubts on whether entailment would be effective in more complex tasks. In this work we show that entailment is also effective in Event Argument Extraction (EAE), reducing the need of manual annotation  to 50\% and 20\% in ACE and WikiEvents respectively, while achieving the same performance as with full training. More importantly, we show that recasting EAE as entailment alleviates the dependency on schemas, which has been a roadblock for transferring annotations between domains. Thanks to the entailment, the multi-source transfer between ACE and WikiEvents further reduces annotation down to 10\% and 5\%  (respectively) of the full training without transfer. Our analysis shows that the key to good results is the use of several entailment datasets to pre-train the entailment model. Similar to previous approaches, our method requires a small amount of effort for manual verbalization: only less than 15 minutes per event argument type is needed, and comparable results can be achieved with users with different level of expertise. 
\end{abstract}

\section{Introduction}

Building Information Extraction (IE) systems for real-world applications is very costly and has suffered from data-scarcity problems, due in part to the expertise and time required to annotate training data at a large scale with sufficient consistency, but also due to poor transfer between domains: IE annotations depend on the schema used in each domain, and moving to new domains requires new schemas, new annotation guidelines and the manual annotation of new data. In many cases, there is some information overlap between schemas, but performing transfer learning to leverage such overlap 
(i.e. learning from \textbf{multiple sources}) can be difficult: it often requires manually mapping labels between schemas, which is typically brittle, cumbersome and requires costly domain expertise \cite{kalfoglou2003ontology}.

In order to save annotation effort, recent work  recasts IE tasks as
Textual Entailment tasks~\cite{white-etal-2017-inference, DBLP:journals/corr/abs-1804-08207, levy-etal-2017-zero, sainz-etal-2021-label}.
For instance, \citet{sainz-etal-2021-label} manually verbalize each relation type in the Relation Extraction (RE) dataset TACRED \cite{zhang-etal-2017-position} to generate hypotheses for each test example, and then apply an entailment model to output the relation type of the hypothesis with highest entailment probability. The entailment model is typically based on large language models pre-trained on entailment datasets such as MNLI \cite{williams-etal-2018-broad}. The approach obtains very strong results on zero-shot and few-shot scenarios, but we note that
TACRED contains relations between two entities that are easily verbalizable,\footnote{For instance, \textsc{per:date\_of\_birth} can be verbalized as \texttt{\{subj\}'s birthday is on \{obj\}} in which subj and obj refers to the two text mentions involved in the relation.} casting doubts on whether entailment would be effective in more complex IE tasks. Event Argument Extraction (EAE) involves more complex contexts, higher ambiguity in the words that trigger events, and depends on the event type in addition to the relation (see Figure \ref{fig:approach}). 

In this work, we present the first system for EAE that addresses the task as an entailment problem. We empirically show the robustness of the method on the zero-shot, few-shot and full training regimes, obtaining state-of-the-art results on ACE \cite{ACE} and WikiEvents \cite{li-etal-2021-document}. 
In addition, we make the following contributions: (1) We show that our method reduces schema dependency, as it improves the performance on the WikiEvents results using additional ACE training data and vice versa with no extra manual work. 
(2) Ablation results show that training with several NLI datasets is significantly better than just using MNLI.
(3) Our analysis of the manual work required for writing templates and annotating arguments sheds light in the sweet spot for future applications, and shows that  template writing does not require much domain expertise
as shown by the results using
an independent novice template writer.
We make the code, templates and models publicly available.\footnote{\url{https://github.com/osainz59/Ask2Transformers}}

\section{Related Work} \label{sec:related_work}

\paragraph{Textual Entailment} Given a textual premise and a hypothesis, the task is to decide whether the premise entails or contradicts (or is neutral to) the  hypothesis \cite{10.1007/11736790_9}. The current state-of-the-art uses large pre-trained Language Models (LM) ~\cite{Lan2020ALBERT:,roberta, conneau-etal-2020-unsupervised, lewis-etal-2020-bart, deberta} fine-tuned on manually annotated datasets such as SNLI \cite{bowman-etal-2015-large}, MNLI ~\cite{williams-etal-2018-broad}, FEVER \cite{thorne-etal-2018-fever} or ANLI \cite{nie-etal-2020-adversarial}. The task is also known as Natural Language Inference (NLI). 

\paragraph{Prompt and Pivot task based learning} has emerged as a candidate solution for data-scarcity problems \cite{le-scao-rush-2021-many,min2021recent,DBLP:journals/corr/abs-2107-13586}. The use of discrete \cite{gao-etal-2021-making, schick-schutze-2021-exploiting, schick-schutze-2021-shot, schick-schutze-2021-just} or continuous \cite{liu2021ptuning} prompts allowed  language models to perform significantly better on many text classification tasks. Closely related to our approach, several works make use of a high-resource supervised task such as Question Answering or entailment as pivot tasks \cite{yin-etal-2019-benchmarking, yin-etal-2020-universal, wang2021entailment, sainz-rigau-2021-ask2transformers, mccann2018natural}.  In the case of entailment, \citet{10.1007/11736790_9} converted QA data to entailment manually and \citet{demszky2018transforming} did it automatically. Other semantic tasks such as Named Entity Recognition, Relation Extraction and Semantic Role Labelling have also been reformulated as entailment by automatically converting data into the entailment format \cite{white-etal-2017-inference, DBLP:journals/corr/abs-1804-08207, levy-etal-2017-zero, sainz-etal-2021-label}. 


\paragraph{Multi-task learning}


reformulates multiple tasks to a single and common task via prompting large pre-trained language models, leveraging multiple data sources to improve each task of interest. Such approaches have shown improvements in supervised \cite{subramanian2018learning, T5, ExT5} and zero-shot scenarios \cite{T0, FLAN}. While using the language modelling task as a pivot shows strong performance with very large language models, it is not clear that smaller models can benefit from this strategy in the same way. \citet{FLAN} and \citet{mishra2021crosstask} obtained contradictory results. In a similar way, Question Answering has been proposed as a pivot task for multi-task learning but without promising results \cite{mccann2018natural}. In this work, we explore multi-source learning, where datasets from different or similar tasks are used to build a model for the target task.

\paragraph{Event Argument Extraction} is a sub-task of Event Extraction. The goal is to identify arguments or fillers for a specific slot (a.k.a., role) in an event template. This task has been largely explored on the Message Understanding Conference (MUC, \citet{grishman-sundheim-1996-message}) and later on Automatic Content Evaluation (ACE). ACE focused mainly on sentence level evaluation due to the difficulty of the task at the time. Recently, new benchmarks such as RAMS \cite{ebner-etal-2020-multi} and WikiEvents have emerged with the aim of addressing document level information extraction similar to MUC. However, most of the interest is still focused on the sentence level.

EAE has been recently addressed by end-to-end event extraction models \cite{wadden-etal-2019-entity,lin-etal-2020-joint,li-etal-2021-future}, instead of treating it as an independent task  \cite{du2020documentlevel}, as we do, or as a subtask in a pipeline \cite{lyu-etal-2021-zero}. 
Lately, with the recent paradigm shift \textbf{to prompt design learning} \cite{min2021recent}, several works reformulated the task as a Question Answering problem \cite{li-etal-2020-event, feng2020probing, du-cardie-2020-event, liu-etal-2020-event, wei-etal-2021-trigger, lyu-etal-2021-zero, sulem-etal-2022-yes} or as a Constrained Text Generation problem \cite{chen-etal-2020-reading, du-etal-2021-grit, li-etal-2021-document} using predefined prompts, questions or templates. We instead reformulate the task as a textual entailment problem. 


\section{Approach}

\begin{figure*}
    \centering
    \resizebox{0.85\textwidth}{!}{
        \includegraphics{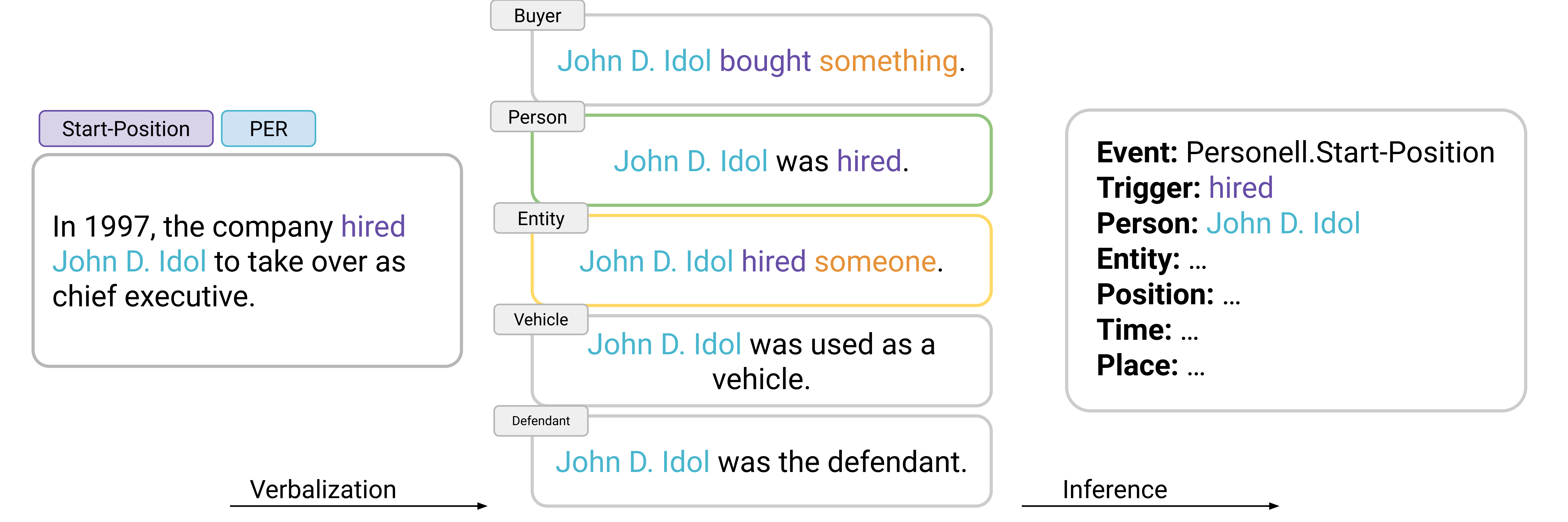}
    }
    \caption{Entailment-based  Event Argument Extraction. On the left, input information: the context, the event trigger ({\color{RoyalPurple} \emph{hired}}) and the argument candidate ({\color{Aquamarine} \emph{John D. Idol}}), alongside the types of both. On the middle, some hypothesis verbalized using the templates: the green box is entailed, the yellow box matches the type constraint but it is not entailed, and the rest do not satisfy type constraints. On the right, the output with the inferred role (Person).}
    \label{fig:approach}
    \vspace{-1em}
\end{figure*}

In order to cast EAE as an entailment task, we verbalize event argument instances using a set of intuitive and linguistically motivated templates to capture the event argument roles, and then perform inferences with entailment models. The entailment model can be additionally trained with EAE training data converted into the entailment format, similar to \citet{sainz-etal-2021-label}. Figure \ref{fig:approach} shows the general workflow of the method. First, the possible roles are verbalized by means of predefined templates and the input, which comprises the context, trigger and argument candidate. Then, an entailment model is used to generate the entailment probability for each verbalization. To predict the role, the most probable hypothesis (verbalization) is chosen among the roles that satisfy the event-entity\footnote{In this context, entities also include values such as time or amounts.} constraints. A more detailed description of each component follows.


\paragraph{Label verbalization}

\begin{table*}
    \centering
    \resizebox{\textwidth}{!}{
        \begin{tabular}{r|p{7cm}l}
            \toprule
            Template type & Description & Example \\
            \midrule
            {\color{Aquamarine} \{arg\}} & Templates with \textbf{implicit} information about the event. {\color{Aquamarine} \{arg\}} variable is the placeholder for the argument candidate. & {The victim was {\color{Aquamarine} \{arg\}}.} \\
            {\color{RoyalPurple} \{trg\}} $\rightarrow$ {\color{Aquamarine} \{arg\}} & Templates with \textbf{explicit} information about the event. The {\color{RoyalPurple} \{trg\}} variable is the placeholder for the event trigger. & {The {\color{RoyalPurple} \{trg\}} occurred in {\color{Aquamarine} \{arg\}}.}\\
            {\color{RoyalPurple} \{canonical(trg)\}} $\rightarrow$ {\color{Aquamarine} \{arg\}} & Templates with predefined canonical values for the {\color{RoyalPurple} \{trg\}} variable. & {{\color{Aquamarine} \{arg\}} was {\color{RoyalPurple} jailed}.} \\
            {\color{RoyalPurple} \{canonical(trg)\}, {\color{Orange} placeholder}} $\rightarrow$ {\color{Aquamarine} \{arg\}} & Templates that makes use of agent or patient dummy placeholders in order to produce grammatical sentences. & {The {\color{Aquamarine} \{arg\}} {\color{RoyalPurple} inspected} {\color{Orange} something}. } \\
            \bottomrule
        \end{tabular}
    }
    \caption{The four main template categories used to create the role verbalizations.}
    \label{tab:template_characterization}
    \vspace{-1em}
\end{table*}

is attained using templates that combine the information of the instance and express a specific label. 
Different role verbalizations are shown in Figure \ref{fig:approach}.
%
%
%
%
%
A verbalization is generated using templates that have been manually written based on the task guidelines of each dataset. The templates involve the candidate argument, and optionally the event trigger. In some cases, in order to produce a grammatical hypothesis, placeholders corresponding to the agent or theme are also introduced, which can be generic, e.g. \emph{someone}, or dependent of the argument role, e.g. \emph{defendant}. 
We defined several template types (see Table \ref{tab:template_characterization}) to guide the creation of templates more systematically. In Section \ref{subsec:verbalization_design} we describe the process to create templates, and in Section \ref{sec:analysis} we analyse the differences between independent template developers and how this did not affect performance. The templates created for the ACE dataset are listed in Appendix \ref{sec:template_devs}.

\paragraph{Entailment model.} Given a premise and hypothesis, the model returns the probabilities of the hypothesis being entailed by, contradicted to or neutral to the premise. In principle, any model trained on the NLI task can be used. 

\paragraph{Inference}
takes into account three key factors to output the role label for an argument candidate: the entailment probabilities of each verbalization, the type constraints of the specific role, and a threshold. Argument candidates which do not match the type constraints are discarded. From the rest, we return the role of the verbalized hypothesis with highest entailment probability, unless the probability is lower than the threshold, in which case we return the negative class.\footnote{The class that represents that the argument candidate takes no part on the event.}

\paragraph{Training.}
Our entailment-based model can be applied without any training on the EAE task, in a zero-shot fashion, or, alternatively, the entailment model can be finetuned using training data from the EAE dataset. 
For this purpose, we convert the EAE training dataset into a NLI format, i.e we generate entailment, neutral and contradiction hypotheses heuristically from the data using the templates themselves. For each positive labeled example (a candidate that is an argument) we sample $N_{E}$ entailment hypotheses using the templates that correspond to the correct label and $N_N$ neutral hypotheses using templates from different roles. For each negative example (the candidate is not an argument of the event) we create $N_C$ contradiction hypotheses using any template at random. $N_E$, $N_N$ and $N_C$ are considered hyperparameters of the training phase along with the hyperparameters of the neural network model such as learning-rate and batch-size.  In order to create challenging training examples for the negative class, we propose to use \textbf{constrained sampling}, based on the trigger-entity type constraints, where we create negative examples from candidates that satisfy the constraints. Preliminary experiments showed slight improvements with respect to regular sampling.

\section{Entailment for Multi-source Learning}

\begin{figure}
    \centering
    \resizebox{\linewidth}{!}{
        \includegraphics{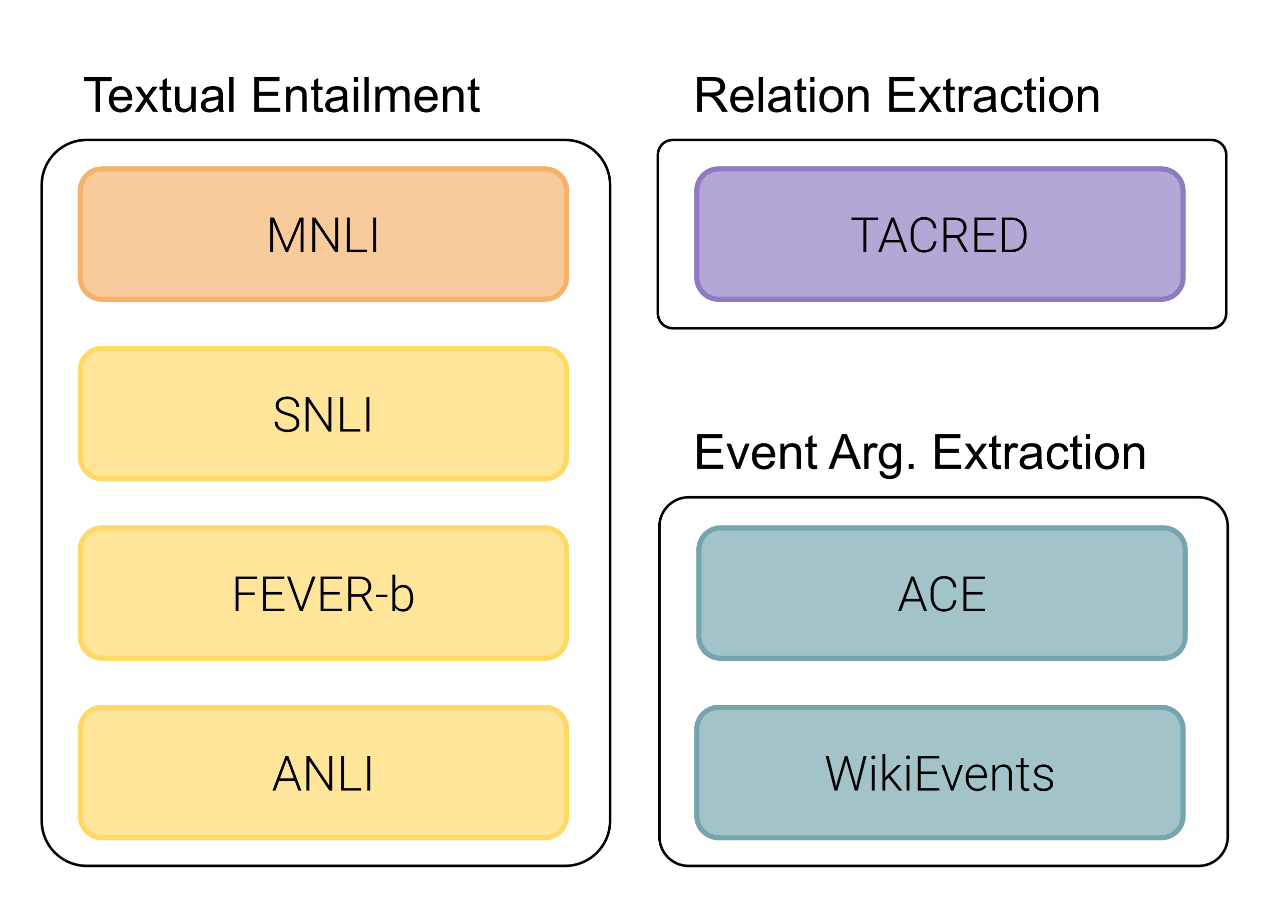}
    }
    \caption{Datasets used by task category.}
    \label{fig:task_taxonomy}
    \vspace{-1em}
\end{figure}


We hypothesize that two similar IE tasks can benefit from each other even if they do not share the same schema or domain. Although this hypothesis is very intuitive and it has been demonstrated on several works for tasks other than IE (see Multi-task learning on Section \ref{sec:related_work}), actual IE models are limited by  schema dependency, which makes it almost impossible to learn from datasets annotated with different IE schemas. One option is to perform a manual mapping between schemas, which is costly and often inaccurate \cite{kalfoglou2003ontology}. 
Our approach instead is domain and schema agnostic, and therefore allows to learning from multiple sources seamlessly.  Given that the sources are recast into a single format in a common entailment formulation, it suffices to fine-tune the model in sequence across the sources.

To check our hypothesis we split tasks according to the following criteria: (1) IE sources like Relation Extraction that are different from EAE (e.g. TACRED), and (2)  EAE sources using different schemas (e.g. WikiEvents and ACE). Figure \ref{fig:task_taxonomy} summarizes the tasks and datasets used in this work, including the four natural language understanding datasets.



\section{Experimental Setup}

In this section, we describe the methodology for template development, evaluation setting, the baselines used in our experiments, and the computation infrastructure specifications.

\subsection{Methodology for verbalization} \label{subsec:verbalization_design}

The templates used to generate the verbalizations were created based on the annotation guidelines of each dataset. During the creation, the template developers had access to the guidelines that describe each of the roles (which can include one or two examples) and a NLI model that the developer could use to verify whether the generated verbalizations of these examples were entailed by the model. The developer was allowed a maximum of 15 minutes per role, and spent 5 and 12 hours\footnote{Given that there is a total of 22 and 59 role types respectively, this is equivalent to an average of 13 and 12 minutes per role.} to create the templates for ACE and WikiEvents respectively. 


\subsection{Evaluation} \label{subsec:evaluation}


\begin{table}
    \centering
    \begin{tabular}{r|rrrr}
        \toprule
         & \multicolumn{2}{c}{ACE} & \multicolumn{2}{c}{WikiEvents} \\
         Train split & \# Pos & Total & \# Pos & Total \\
        \midrule
        0\% & - & - & - & - \\
        1\% & 2.05 & 173 & 0.86 & 195 \\
        5\% & 11.36 & 843 & 4.09 & 966 \\
        10\% & 23.86 & 1736 & 8.26 & 1903 \\
        20\% & 45.00 & 3302 &  15.84 & 3578 \\
        100\% & 220.86 & 16502 & 79.68 & 18532 \\
        \bottomrule
    \end{tabular}
    \caption{Mean examples per role (pos) and total number of examples (positive and negative) across different training data splits and datasets.}
    \label{tab:train_splits}
    \vspace{-1em}
\end{table}

\paragraph{Datasets.} We carried out our evaluation on two different EAE datasets: ACE \cite{ACE} and WikiEvents \cite{li-etal-2021-document}. The ACE2005 dataset is a sentence-level Event Extraction dataset that contains entities, relations, event-triggers and arguments annotations on English, Chinese and Arabic texts. We worked only on the English EAE task. The WikiEvents dataset is instead more focused on document-level argument extraction task. Although the last is intended to be use as a document-level benchmark we focused on the sentence-level extraction\footnote{We consider as model prediction errors the arguments that are outside the sentence, to be consistent with other systems evaluation.} for two reasons: to maintain consistency with ACE dataset and because the nearest occurrence of the arguments are inside the sentence of the event trigger in almost all examples. For both ACE and WikiEvents, we split the training data into different amounts (0\%, 1\%, 5\%, 10\%, 20\% and 100\%) following \citet{liu-etal-2020-event} to also evaluate our system on extreme data scarcity scenarios. Table \ref{tab:train_splits} shows the amount of examples per split. The total amount refers to the addition of all positives and negatives trigger-candidate  pairs. 

\paragraph{Metrics.} We have used the standard F1-Score, which is a common metric on IE tasks. Along with that, we propose the use of the Area Under the Curve (AUC) for better model comparison across all scenarios. The reported AUC scores are computed with all splits for the main results and just with 0\%, 5\% and 100\% for the multi-source results, and therefore, they are not comparable. 

\subsection{Baselines and Models}

\paragraph{Baselines.} Our main point of comparison is our re-implementation of EM 
\cite{baldini-soares-etal-2019-matching}, as we can run it on the same few-shot splits as our system and allow for head-to-head comparison. EM is a state-of-the-art~\cite{https://doi.org/10.48550/arxiv.2102.01373} model that uses \textsc{RoBERTa\textsubscript{large}} as a backbone. In addition we  
also report results of the state-of-the-art models that have been run on our same experimental setup, having access to gold event-trigger and entity annotations. On ACE, we report the results of BERTEE and RCEE\_ER, both reported at \cite{liu-etal-2020-event}, which correspond to a BERT \cite{devlin-etal-2019-bert} based baseline and a QA based pivot approach that leverages SQuAD \cite{rajpurkar-etal-2016-squad} data. Unfortunately 
the data splits used by \cite{liu-etal-2020-event} are not available\footnote{Personal communication.} and thus, only the results for zero-shot (i.e. 0\% training data) and full training (i.e. 100\% training data) are directly comparable.
Regarding WikiEvents Gen-Arg \cite{li-etal-2021-document} uses gold triggers, but not gold entity information, so we decided to report Coref-F1\footnote{We used this to alleviate the noise introduced by not using the gold entity annotations, and therefore, make the comparison more fair.} which refers to the F1-Score of predicting at least one of the gold entity coreferential chain as argument. 

\paragraph{NLI models} used in this work are based on the RoBERTa\textsubscript{large} \cite{roberta} checkpoint, and are available via HuggingFace Transformer's model repository \cite{wolf-etal-2020-transformers}. The main results use a model trained on all MNLI, SNLI, FEVER and ANLI, and in the analysis we also report the results of a model using just MNLI (see Appendix \ref{ap:hyperparameters} for more information, including hyperparameters used).

\subsection{Infrastructure}
All the experiments were done in a \textbf{single} RTX 2080ti (11Gb) with a 250W power consumption. The average training times are:\footnote{The time required for training the model depends linearly with the sampling rates of entailment, neutral and contradiction examples.} 0.36h/epoch for ACE, 0.52h/epoch for WikiEvents and 2.86 h/epoch for TACRED. In total, 464.56 hours (154.86 if only a single run is done) of computation time are required to reproduce \textbf{all} the experiments, that in our setting corresponds to 21.36 kgCO$_2$eq carbon footprint\footnote{Estimation based on \url{mlco2.github.io/impact/}} (roughly equivalent to the CO\textsubscript{2} emitted by 88.2 km driven by an average car).

\begin{table*}[t]
    \centering
        \begin{tabular}{llllllll}
            \toprule
             \multicolumn{8}{c}{ACE} \\
            Model & 0\% & 1\% & 5\% & 10\% & 20\% & 100\% & AUC\\
            \midrule
            BERTEE  & \multicolumn{1}{c}{-} & *2.20 & *10.5 & *19.3 & *28.6 & 64.7 & *40.73 \\ 
            EM & \multicolumn{1}{c}{-} & 4.58 \small{$\pm 1.55$} & 37.5 \small{$\pm 2.98$} & 50.9 \small{$\pm 0.96$} & 58.7 \small{$\pm 1.9$} & 72.1 \small{$\pm 0.65$} & 60.87 \\ 
            RCEE\_ER  & 37.0 & *49.8 & *59.9 & *65.1 & *67.6 & 70.1 & *67.47 \\ 
            \midrule
            NLI & \textbf{40.6} & \textbf{45.4} \small{$\pm 0.16$} & \textbf{57.1} \small{$\pm 0.93$} & \textbf{64.6 \small{$\pm 1.12$}} & \textbf{69.8 \small{$\pm 0.58$}} & \textbf{74.6 \small{$\pm 0.88$}} & \textbf{70.00} \\ 
            \midrule
              \multicolumn{8}{c}{WikiEvents} \\
            Model & 0\% & 1\% & 5\% & 10\% & 20\% & 100\% & AUC\\
            \midrule
            EM & \multicolumn{1}{c}{-} & 16.9 \small{$\pm 0.63$} & 41.5 \small{$\pm 1.47$} & 49.9 \small{$\pm 0.28$} & 54.9 \small{$\pm 1.30$} & 61.3 \small{$\pm 1.04$} & 55.26 \\ 
            *Gen-Arg & \multicolumn{1}{c}{-} & 2.4 \small{$\pm 1.66$} & 30.5 \small{$\pm 4.12$} & 48.1 \small{$\pm 1.42$} & 55.7 \small{$\pm 1.35$} & 65.1 & 56.15 \\ 
            \midrule
            NLI & \textbf{35.9} & \textbf{42.6 \small{$\pm 1.36$}} & \textbf{52.2 \small{$\pm 1.40$}} & \textbf{59.5 \small{$\pm 0.58$}} & \textbf{65.4 \small{$\pm 0.62$}} & \textbf{69.9 \small{$\pm 0.70$}} & \textbf{65.45} \\ 
            \bottomrule
        \end{tabular}
    \caption{Main results on different training data splits for our NLI model, EM baseline and state-of-the-art systems. 
    * for results not directly comparable with ours. Bold for best among comparable results.}
    \label{tab:main_results}
\end{table*}

\begin{table*}[t]
    \centering
    \resizebox{\textwidth}{!}{
        \begin{tabular}{l|llll|llll}
            \toprule
             & \multicolumn{4}{c|}{ACE} & \multicolumn{4}{c}{WikiEvents} \\
            Source & 0\% & 5\% & 100\% & AUC & 0\% & 5\% & 100\% & AUC\\
            \midrule
            NLI & 40.6 & 57.1 \small{$\pm 0.93$} & 74.6 \small{$\pm 0.88$} & 65.0 & 35.9 & 52.2 \small{$\pm 1.40$} & 69.9 \small{$\pm 0.70$} & 60.2\\
            \midrule
            NLI \textsubscript{+ WikiEvents} & \textbf{62.7} & \textbf{69.3} \small{$\pm 0.35$} & \textbf{74.9} \small{$\pm 0.58$} & \textbf{71.8} & \multicolumn{1}{c}{-} & \multicolumn{1}{c}{-} & \multicolumn{1}{c}{-} & \multicolumn{1}{c}{-} \\
            NLI \textsubscript{+ ACE} & \multicolumn{1}{c}{-} & \multicolumn{1}{c}{-} & \multicolumn{1}{c}{-} & \multicolumn{1}{c|}{-} & \textbf{57.3} & 65.2 \small{$\pm 0.41$} & \textbf{71.5} \small{$\pm 1.07$} & \textbf{68.0} \\
            \midrule
            NLI \textsubscript{+ RE} & 44.5 & 56.3 \small{$\pm 0.79$} & 73.9 \small{$\pm 0.05$} & 64.4 & 38.2 & 55.0 \small{$\pm 1.38$} & 69.2 \small{$\pm 0.59$} & 61.3 \\
            NLI \textsubscript{+ RE + WikiEvents} & \textbf{62.7} & 65.9 \small{$\pm 0.30$} & 74.0 \small{$\pm 0.49$} & 69.7 & \multicolumn{1}{c}{-} & \multicolumn{1}{c}{-} & \multicolumn{1}{c}{-} & \multicolumn{1}{c}{-} \\
            NLI \textsubscript{+ RE + ACE} & \multicolumn{1}{c}{-} & \multicolumn{1}{c}{-} &\multicolumn{1}{c}{-} & \multicolumn{1}{c|}{-} & 56.7 & \textbf{66.4 \small{$\pm 0.95$}} & 69.8 \small{$\pm 2.68$} & 67.8 \\
            \bottomrule
        \end{tabular}
    }
    \caption{Multi-source learning results of the NLI model. The AUC score reported on this table is only computed with 0\%, 5\% and 100\% points, and therefore, is not comparable with Table \ref{tab:main_results}. RE is shorthand for TACRED.}
    \label{tab:multi_task_results}
    \vspace{-0.75em}
\end{table*}

\section{Results}


\paragraph{Main results.}

Table \ref{tab:main_results} reports our NLI system, including the median F1-Score and the standard deviation across 3 different runs 
of our implementations NLI and EM. On ACE our system is best on all comparable results and overall as shown by the AUC score. 
On the case of WikiEvents, our system is the best in all cases. In both datasets the EM baseline is outperformed by the NLI system. 

\paragraph{Multi-source results.}

Table \ref{tab:multi_task_results} describes our multi-source learning results, where we use  NLI+ to indicate systems that use additional sources for training. We report the median F1-Score across 3 runs for 0\%, 5\% and 100\% scenarios and the corresponding AUC score on ACE and WikiEvents. The rows show the impact of transferring knowledge from the training part of different tasks (for more detailed per role analysis see Appendix \ref{sec:multi-task-in-depth}). The results show that the signal between EAE datasets (i.e. WikiEvents and ACE) is strong, yielding significant improvements in all scenarios. For instance, on zero-shot evaluation, the systems obtain the impressive scores of 62.7 and 57.3, close to 20 points of improvement.

Sequentially fine-tuning our NLI model in TACRED and then in our target task shows small improvements on low-resource scenarios (0\% split for ACE, 0\% and 5\% splits for WikiEvents). 
Training on the three sources sequentially does not seem to yield further improvements. 

\begin{figure}[t]
    \centering
    \resizebox{\linewidth}{!}{
        \includegraphics{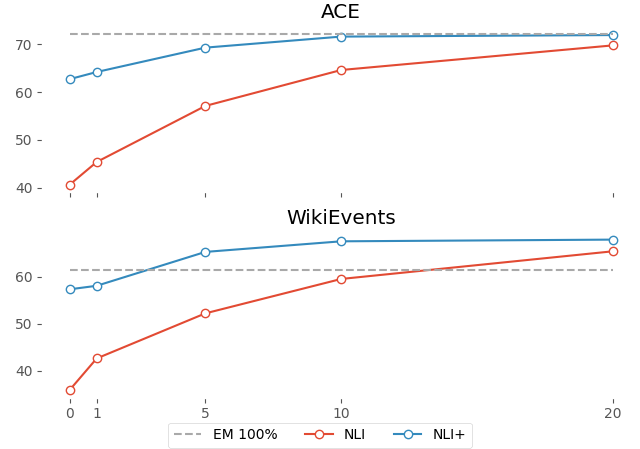}
    }
    \caption{Comparison between the baseline EM model trained on 100\% training, and our NLI and multi-source enhanced NLI+ models (NLI\textsubscript{+ WikiEvents} and NLI\textsubscript{+ ACE}) with different training subsets.}
    \label{fig:multi_task_results}
    \vspace{-1em}
\end{figure}

Figure \ref{fig:multi_task_results} shows the performance of our NLI and multi-source enhanced NLI+ systems along with the EM 
baseline (data from Tables \ref{tab:main_results} and \ref{tab:multi_task_results}). The curves show that our NLI+ systems only need 10\% and 5\% of the data (on ACE and WikiEvents, respectively) to outperform the EM baseline that uses 100\% of the training data.


\section{Analysis} \label{sec:analysis}

After performing the main experiments we did some additional analysis.

\paragraph{The importance of using several NLI datasets.}

A perfect NLI model should, in theory, solve any task that is framed correctly as entailment. Of course, there is not "perfect" NLI model. In fact, current state-of-the-art NLI models tend to learn artifacts and lexical patterns \cite{gururangan-etal-2018-annotation, poliak-etal-2018-hypothesis, tsuchiya-2018-performance, glockner-etal-2018-breaking, geva-etal-2019-modeling, mccoy-etal-2019-right} instead of the task itself. Motivated by these issues, datasets like ANLI \cite{nie-etal-2020-adversarial} were adversarially created to alleviate them. The lack of robustness of NLI models gets amplified when it comes to a cross-task evaluation. For instance, the model trained on MNLI achieves 90.2 accuracy on MNLI and 31.4, 29.5 and 55.6 F1-Score on ACE, WikiEvents and TACRED respectively (cf. Table \ref{tab:re_results}). Adding FEVER, SNLI and ANLI to the training improves MNLI accuracy only 0.8 points to 91.0, but zero-shot scores on ACE, WikiEvents and TACRED improve +9.2, +6.4 and +1.2 respectively. In few-shot and full-training scenarios, the results also improve when using several NLI datasets. 
Our results suggest that new, more challenging NLI datasets, as well as NLI datasets automatically generated from other sources (as done in this work with WikiEvents and ACE) will yield  more robust entailment models, and could further increase the performance of entailment-based EAE and IE.

\begin{table}[t]
    \centering
        \resizebox{\linewidth}{!}{
        \begin{tabular}{l|llll}
            \toprule
            Model \textsubscript{source} & 0\% & 5\% & 100\% & AUC \\
            \midrule
            \multicolumn{5}{c}{ACE} \\
            \midrule
            NLI \textsubscript{MNLI only} & 31.4 & 46.0 \small{$\pm 0.55$} & 62.8 \small{$\pm 2.83$} & 53.6 \\
            NLI & \textbf{40.6} & \textbf{57.1} \small{$\pm 0.93$} & \textbf{74.6} \small{$\pm 0.88$} & \textbf{65.0} \\
            \midrule
            \multicolumn{5}{c}{WikiEvents} \\
            \midrule
            NLI \textsubscript{MNLI only} & 29.5 & 49.3 \small{$\pm 0.32$} & 59.9 \small{$\pm 0.99$} & 53.8 \\
            NLI & \textbf{35.9} & \textbf{52.2} \small{$\pm 1.40$} & \textbf{69.9} \small{$\pm 0.70$} & \textbf{60.2} \\
            \midrule
            \multicolumn{5}{c}{TACRED} \\
            \midrule
            NLI \textsubscript{MNLI only} & 55.6 & 64.1\small{$\pm 0.20$} & 71.0 & 67.2 \\
            NLI & \textbf{56.8} & \textbf{70.5}\small{$\pm 0.62$} & \textbf{73.2}\small{$\pm 0.65$} & \textbf{71.4} \\
            \bottomrule
        \end{tabular}
    }
    \caption{Ablation on NLI datasets used to-pretrain our NLI model on three datasets. NLI for our system using MNLI, FEVER, SNLI and ANLI (taken Table \ref{tab:main_results}) and NLI \textsubscript{MNLI only} for our system when using MNLI only.}
    \label{tab:re_results}
    \vspace{-1em}
\end{table}

\paragraph{The impact of different template developers.}  \label{subsec:annotators_comparison}
In order to test the robustness of the templates, we enrolled a linguist with experience in NLP annotation but no prior contact with the project nor access to the original templates from the main developer. Under the same time and resource conditions, she was asked to write templates for the ACE dataset. The templates written by the main developer and the linguist vary in different ways: (1) the number of created templates per role and (2) the verbalization style, as the main developer tended to use finite and conjugated verbs while the linguist tended to use infinitives and lemmas. The templates of both are available in Appendix \ref{sec:template_devs}.

\begin{figure}[t]
    \vspace{-0.5em}
    \centering
    \resizebox{\linewidth}{!}{
        \includegraphics{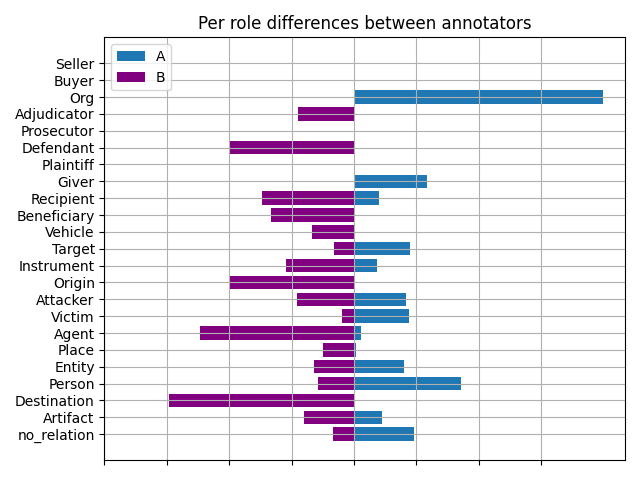}
    }
    \caption{Recall differences between the main developer (A, right) and the linguist (B, left).}
    \label{fig:annotators_differences}
    \vspace{-1em}
\end{figure}

To study the performance of the templates of each developer per role,  
Figure \ref{fig:annotators_differences} shows the instances that a system correctly classified and the other system did not, and vice versa. The bars display the recall, as they are normalized by the frequencies of the roles. Missing bars on a row means that both performed the same on that role (e.g. Seller). When only a blue bar is shown (e.g. Org) it means that the main developer recovered arguments which the linguist did not, \textbf{and} there were no examples where the linguist recovered arguments that the developer did not. The same applies to situations where there is only purple bars. Roles with mixed results include examples where one or the other succeeded. As we can see, the approaches seem to be complementary, with the linguist having a higher recall with the roles that are more associated with classical semantic roles. 
Table \ref{tab:annotators_comparison} shows that in general, the templates of the linguist perform similarly to those of the main developer, except for 100\% of the data, where the templates of the main developer were slightly better. 


\begin{table}
    \centering
    \resizebox{\linewidth}{!}{
        \begin{tabular}{l|llllll}
            \toprule
            Developer & 0\% & 1\% & 5\% & 10\% & 20\% & 100\% \\
            \midrule
            (A) Main  & 40.3 & 46.2 & 56.3 & 63.8 & 69.6 & 76.4 \\
            (B) Linguist & 40.4 & 44.9 & 57.3 & 64.2 & 70.1 & 73.3 \\
            \midrule
            $\Delta$ F1 & -0.1 & +1.3 & -1.0 & -0.4 & -0.5 & +3.1 \\
            \bottomrule
        \end{tabular}
    }
    \caption{Results for templates from two developers. Median F1 on the development set are reported.}
    \label{tab:annotators_comparison}
    \vspace{-1em}
\end{table}

%
%
%
%

\paragraph{Verbalizations vs. annotations}

Finally, we carried out an experiment to compare the time and effort requirements of annotation vs. writing the templates. To that end, the linguist re-annotated a small portion of ACE with the same information she had as she was creating the templates. That is, given the argument candidates for each event trigger in the document, she needs to decide whether the candidate was an argument and the type of the argument. She has access to the guidelines (similar to creating the templates), though she did not study them beforehand. Note also that she did the annotations \textbf{after} writing the templates, so she was already familiar with the slots.
Under these conditions, she annotated  46 pairs (event trigger, potential argument candidate) in 30 minutes. Taking into account that  ACE has 16.5000 such pairs, it would take approximately 180 hours to annotate ACE training part. Note that in practice, ACE requires much more time than our estimate to achieve the desired level of quality:
the ACE annotation procedure involved double annotation and a second pass with a senior annotator \cite{doddington2004automatic}. For an analysis of the annotation procedure the interested reader is referred to \citet{min2012compensating}. 

\begin{figure}
    \centering
    \resizebox{\linewidth}{!}{
        \includegraphics{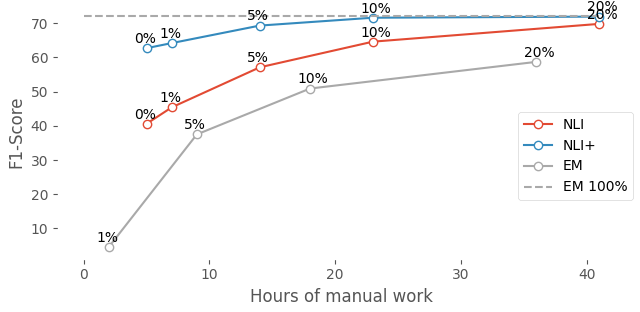}
    }
    \caption{Performance on ACE according to our estimations of manual work in hours. We also indicate the percentage of training data used. }
    \label{fig:results_per_time}
    \vspace{-1em}
\end{figure}

Based on our estimation, 9 hours would allow an annotator to annotate 5\% of the dataset which yields a 37.5 F1  (Figure \ref{fig:results_per_time}),
while 5 hours of template building yields 40.6 F1-Score in the zero-shot setting. With 18 hours 10\% would be annotated and the F1-Score will be 50.9, while 5 hours of template building and 9 hours of annotations would yield 57. Figure \ref{fig:results_per_time} plots the performance according to manual hours on ACE, showing the huge gains provided by the initial 5 hours writing templates, plus the reuse of WikiEvents annotations.
According to our experience, more hours on template building does not necessarily lead to improvements (contrary to annotation), so a \textbf{sweet spot for time investment} seems to be to firstly create templates, and then spend the remaining budget on annotating examples. 

On another note, the linguist mentioned that writing templates is more natural and rewarding than annotating examples, which is more repetitive, stressful and tiresome. When writing templates, she was thinking in an abstract manner,  trying to find generalizations, while  she was paying attention to concrete cases when doing annotation. 

\section{Conclusions}

This paper shows the entailment-based approach for event argument extraction is extremely effective in zero-shot, few-shot and full train scenarios both on ACE and WikiEvents, outperforming previous methods. First of all, recasting EAE as an entailment task allows it to reuse annotations from different event schemas, achieving large gains when transferring annotations between ACE and WikiEvents, and also some gains in the zero-shot performance when transferring annotations from a relation extraction model such as TACRED. Secondly, we show that using additional training entailment datasets improves results significantly over just using MNLI, not only on EAE but also on TACRED. Thirdly, we show that the relatively short time spent writing manual templates is much more effective than the time spent on doing annotations, with a sweet spot where the annotation effort is split between the two, with large savings in manual labour. Lastly, we show that an independent linguist is able to write templates with comparable performance without any special training. We think that our results and analysis support the potential of entailment models for other NLP tasks.

Our work paves the way for a new paradigm for IE, where the expert defines the schema using natural language and directly runs those specifications, annotating a handful of examples in the process, and allowing for quick trial-and-error iterations. \citet{sainz-etal-2021-zs4ie} propose a user interface alongside this paradigm. More generally, inference capability could be extended, acquired and applied from other tasks, in a research avenue where entailment and task performance improve in tandem.  

\section*{Acknowledgements}
Oscar is funded by a PhD grant from the Basque Government (PRE\_2020\_1\_0246). This work is based upon work partially supported via the IARPA  BETTER Program contract No. 2019-19051600006 (ODNI, IARPA), and by the Basque Government (IXA excellence research group IT1343-19).

\typeout{}
\bibliography{anthology,custom}
\bibliographystyle{acl_natbib}

\clearpage

\appendix

\begin{figure*}
    \centering
    \resizebox{\textwidth}{!}{
        \includegraphics{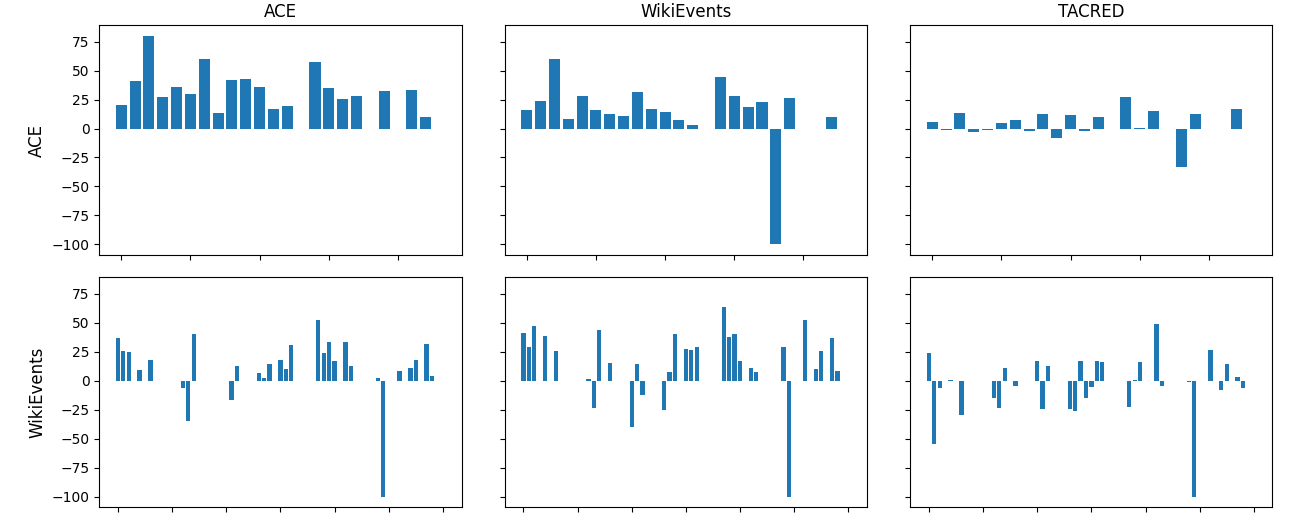}
    }
    \caption{Absolute improvements over the NLI baseline using different tasks and sources. Rows indicates the testing data and columns the training data. Each bar indicates the F1-Score difference between the trained NLI system vs 0\% NLI for a specific role.}
    \label{fig:multi-source/task_contributions}
\end{figure*}

\section{Hyperparameters}
\label{ap:hyperparameters}

\begin{table}
    \centering
    \resizebox{\linewidth}{!}{
    \begin{tabular}{l|ccc}
        \toprule
        Hyperparameter & EM & NLI & NLI\textsubscript{MNLI only} \\
        \midrule
        N\textsubscript{E} / N\textsubscript{N} / N\textsubscript{C} & - & 2 / 5 / 5 & 2 / 5 / 5 \\ 
        Batch size & \multicolumn{3}{c}{32} \\
        Learning rate & $1 \times 10^{-5}$ & $4 \times 10^{-6}$ & $1 \times 10^{-5}$ \\
        Seeds & \multicolumn{3}{c}{\{0, 24, 42\}} \\
        Epochs & \multicolumn{3}{c}{25 (*50)} \\
        Weight decay & \multicolumn{3}{c}{0.01} \\
        \bottomrule
    \end{tabular}
    }
    \caption{Hyperparameters of the trained systems. * indicates the difference between full-train and few-shot scenarios.}
    \label{tab:hyperparameters}
\end{table}

On this section we describe the hyperparameters we have used on our experiments. All the hyperparameters optimized on this work were optimized for the 100\% split with the batch-size fixed to 32, and used on the rest. The Table \ref{tab:hyperparameters} describes the hyperparameters used on EM, NLI and NLI\textsubscript{MNLI only} variants, for the NLI+ the same hyperparameters as NLI were used. We have found that the same exact hyperparameters were the best on ACE, WikiEvents and TACRED datasets. For the future, we plan to test new hyperparameter sets that uses bigger batch-sizes, as recent works \cite{ExT5} suggest to be optimal for multi-task and -source learning experiments.

The pre-trained NLI models used on this work can be downloaded from the HuggingFace Models repository: NLI \textsubscript{MNLI only} (\path{roberta-large-mnli}) and NLI (\path{ynie/roberta-large-snli_mnli_fever_anli_R1_R2_R3-nli}).

The fine-tuned models derived from this work will be uploaded to HuggingFace Models repository. Check the GitHub repository for updated information.

\section{Multi-task in-depth analysis} \label{sec:multi-task-in-depth}

The Figure \ref{fig:multi-source/task_contributions} shows the per role absolute improvement obtained by training on different tasks over the 0\% NLI system. Overall, we can see that training on ACE or WikiEvents improves almost all the roles and training on TACRED improves some and some others do not. A result that was unexpected is that there are few roles on WikiEvents that after training on WikiEvents become worse in contrary to training on ACE. This could be explained by the differences among the frequency distributions that the train, development and test sets of WikiEvents has. Moreover, there are some roles on WikiEvents that decreases in all training scenarios, this suggests us that sequential fine-tuning might be not the best option for this type of multi-source learning and therefore further ways should be explored.

\section{ACE templates from both developers} \label{sec:template_devs}
The next table contains the templates written by both developers for the ACE arguments. We follow the notation introduced in Section \ref{subsec:verbalization_design}. In addition, we also consider information from the event, such as the type on different granularity levels, including \texttt{\{trg\_type\}} for the trigger type (e.g. \textit{Movement} from \textit{Movement.Transport}) and \texttt{\{trg\_subtype\}} for the subtype of the trigger, e.g. \textit{Transport} from \textit{Movement.Transport}).

\clearpage
\onecolumn

\begin{longtable*}{rll} 
    \toprule
    Role & Main developer  & Linguist \\
    \midrule
    \endfirsthead
    \toprule
    Role & Main developer  & Linguist \\
    \midrule
    \endhead
    \multicolumn{3}{c}{(continued on the next page)}
    \endfoot
    \endlastfoot

Adjudicator &  {\color{Aquamarine} \{arg\}} {\color{RoyalPurple} tried} {\color{Orange}the defendant}.  &  {\color{Aquamarine} \{arg\}} {\color{RoyalPurple} convict} {\color{Orange} someone}. \\
            &  {\color{Aquamarine} \{arg\}} {\color{RoyalPurple} convicted} {\color{Orange}the defendant}.  & {\color{Aquamarine} \{arg\}} {\color{RoyalPurple} sentence} {\color{Orange} someone}. \\ 
            &  {\color{Aquamarine} \{arg\}} {\color{RoyalPurple} released} {\color{Orange}the defendant}. &  {\color{Aquamarine} \{arg\}} {\color{RoyalPurple} judge} {\color{Orange} someone}. \\ 
            &  {\color{Aquamarine} \{arg\}} {\color{RoyalPurple} sentenced} {\color{Orange}the defendant}. &  {\color{Aquamarine} \{arg\}} {\color{RoyalPurple} fine} {\color{Orange} someone}. \\ 
            &  {\color{Aquamarine} \{arg\}} {\color{RoyalPurple} acquitted} {\color{Orange}the defendant}. &  {\color{Aquamarine} \{arg\}}  {\color{RoyalPurple} indict} {\color{Orange} someone}. \\  \midrule

Agent & {\color{Aquamarine} \{arg\}} {\color{RoyalPurple} \{trg\}} a {\color{Orange}person or organization}. & {\color{Aquamarine} \{arg\}} {\color{RoyalPurple} do} {\color{Orange} something}. \\ 
      &     & {\color{Aquamarine} \{arg\}} {\color{RoyalPurple} select} {\color{Orange} something}.\\
      &     & {\color{Aquamarine} \{arg\}} {\color{RoyalPurple} carry out} {\color{Orange} something}.\\
      &     & {\color{Aquamarine} \{arg\}} {\color{RoyalPurple} create} {\color{Orange} something}.\\
      &     & {\color{Aquamarine} \{arg\}} {\color{RoyalPurple} give} {\color{Orange} something}. \\ \midrule

Artifact & {\color{Orange} Someone} {\color{RoyalPurple} \{trg\}} the {\color{Aquamarine} \{arg\}}. &  {\color{Aquamarine} \{arg\}} be an object.  \\
         & {\color{Orange} Someone} {\color{RoyalPurple} moved} {\color{Aquamarine} \{arg\}}.  &  {\color{Aquamarine} \{arg\}} be a weapon.    \\
         & {\color{Orange} Someone} {\color{RoyalPurple} bought} {\color{Aquamarine} \{arg\}}.  &    \\
         & {\color{Orange} Someone} {\color{RoyalPurple} sold} {\color{Aquamarine} \{arg\}}.  &   \\ \midrule

Attacker & {\color{Aquamarine} \{arg\}} {\color{RoyalPurple} \{trg\}} a {\color{Orange}person or organization}. & {\color{Aquamarine} \{arg\}} {\color{RoyalPurple}assail} {\color{Orange} someone}.\\
         &   & {\color{Aquamarine} \{arg\}} {\color{RoyalPurple}aggress} {\color{Orange} someone}.\\
         &   & {\color{Aquamarine} \{arg\}} {\color{RoyalPurple}assault} {\color{Orange} someone}.\\ \midrule

Beneficiary & {\color{Orange}The buyer} {\color{RoyalPurple} bought} to {\color{Aquamarine} \{arg\}} {\color{Orange} something}.
 & {\color{Aquamarine} \{arg\}} {\color{RoyalPurple}get} {\color{Orange} something} .\\
&  & {\color{Aquamarine} \{arg\}} be beneficiary.\\
&  & {\color{Aquamarine} \{arg\}} {\color{RoyalPurple}benefit from} {\color{Orange} something}.\\
&  & {\color{Aquamarine} \{arg\}} {\color{RoyalPurple}obtain} {\color{Orange} something}.\\ \midrule

Buyer & {\color{Aquamarine} \{arg\}} {\color{RoyalPurple} bought} {\color{Orange} something}. & {\color{Aquamarine} \{arg\}} {\color{RoyalPurple}buy} {\color{Orange} something}. \\
      &  & {\color{Aquamarine} \{arg\}} {\color{RoyalPurple}possess} {\color{Orange} something}. \\
      &  & {\color{Aquamarine} \{arg\}} {\color{RoyalPurple}own} {\color{Orange} something}.\\ \midrule    

Defendant&  {\color{Aquamarine} \{arg\}} was the defendant. & {\color{Aquamarine} \{arg\}} be {\color{RoyalPurple}accused} of {\color{Orange} something}. \\
&  & {\color{Aquamarine} \{arg\}} be {\color{RoyalPurple}accused} of {\color{Orange}a crime}.\\
&  & {\color{Aquamarine} \{arg\}} be {\color{RoyalPurple}judged}. \\ \midrule        
        
Destination & {\color{Orange} Someone} {\color{RoyalPurple} \{trg\_subtype\}} to {\color{Aquamarine} \{arg\}}. & {\color{RoyalPurple} \{trg\_type\}} go to {\color{Aquamarine} \{arg\}}. \\
   &  & {\color{RoyalPurple} \{trg\_type\}} finish in {\color{Aquamarine} \{arg\}}. \\
&  & {\color{RoyalPurple} \{trg\_type\}} move to {\color{Aquamarine} \{arg\}}. \\
&  & {\color{Aquamarine} \{arg\}} be a place. \\
&  & {\color{Aquamarine} \{arg\}} be a location. \\ \midrule   

Entity & {\color{Aquamarine} \{arg\}} {\color{RoyalPurple} attended} {\color{Orange}the demonstration}. & {\color{Aquamarine} \{arg\}} {\color{RoyalPurple}select} {\color{Orange} something}. \\
&  {\color{Aquamarine} \{arg\}} {\color{RoyalPurple} met} {\color{Orange} someone}. & {\color{Aquamarine} \{arg\}} {\color{RoyalPurple}carry out} {\color{Orange} something}.\\
& {\color{Aquamarine} \{arg\}} {\color{RoyalPurple} fired} {\color{Orange} someone}.  & {\color{Aquamarine} \{arg\}} {\color{RoyalPurple}do} {\color{Orange} something}.\\
&  {\color{Aquamarine} \{arg\}} voted in the elections. & {\color{Aquamarine} \{arg\}} {\color{RoyalPurple}create} {\color{Orange} something}.\\
&  {\color{Aquamarine} \{arg\}} {\color{RoyalPurple} released} {\color{Orange}the defendant}. & {\color{Aquamarine} \{arg\}} {\color{RoyalPurple}give} {\color{Orange} something}.\\
& {\color{Aquamarine} \{arg\}} was ordered to pay. &  \\ \midrule
     
Giver &  {\color{Aquamarine} \{arg\}} {\color{RoyalPurple} gave} {\color{Orange} something} to {\color{Orange} someone}.      & {\color{Aquamarine} \{arg\}}  {\color{RoyalPurple}give} {\color{Orange} something}. \\ \midrule
 Instrument &  {\color{Orange} Someone} {\color{RoyalPurple} \{trg\_subtype\}} with {\color{Aquamarine} \{arg\}}. &   {\color{Aquamarine} \{arg\}} be artifact. \\
&  & {\color{Aquamarine} \{arg\}} be object. \\
&  & {\color{Aquamarine} \{arg\}} be device. \\
&  & {\color{Aquamarine} \{arg\}} cause harm. \\ \midrule

Org & {\color{Aquamarine} \{arg\}} organization {\color{RoyalPurple}declared bankruptcy}. & {\color{Aquamarine} \{arg\}} be {\color{RoyalPurple}in bankruptcy.} \\
&   {\color{Aquamarine} \{arg\}} organization was {\color{RoyalPurple} dissolved}. & {\color{Aquamarine} \{arg\}} be {\color{RoyalPurple}ended.} \\ 
&   {\color{Aquamarine} \{arg\}} organization was {\color{RoyalPurple}merged}. & {\color{Aquamarine} \{arg\}} be {\color{RoyalPurple}merged.} \\
& {\color{Aquamarine} \{arg\}} organization was {\color{RoyalPurple}launched}. & {\color{Aquamarine} \{arg\}} be {\color{RoyalPurple}created.} \\
&  & {\color{Aquamarine} \{arg\}} be company. \\
&  & {\color{Aquamarine} \{arg\}} be organization. \\  \midrule       

Origin &  {\color{Orange} Someone} {\color{RoyalPurple} \{trg\_subtype\}} from {\color{Aquamarine} \{arg\}}. & {\color{Aquamarine} \{arg\}} change location. \\
& & {\color{Aquamarine} \{arg\}} be location. \\ 
& &  {\color{RoyalPurple} \{trg\_type\}} start in {\color{Aquamarine} \{arg\}}. \\ 
& & {\color{RoyalPurple} \{trg\_type\}} move from {\color{Aquamarine} \{arg\}} .\\  \midrule       

Person &  {\color{Aquamarine} \{arg\}} was {\color{RoyalPurple} \{trg\}}. & {\color{Aquamarine} \{arg\}} be person. \\
& & {\color{Aquamarine} \{arg\}} be living entity. \\
& & {\color{Aquamarine} \{arg\}} be {\color{RoyalPurple}born.} \\
& & {\color{Aquamarine} \{arg\}} {\color{RoyalPurple} get married.} \\
& & {\color{Aquamarine} \{arg\}} be {\color{RoyalPurple}married.} \\
& & {\color{Aquamarine} \{arg\}} {\color{RoyalPurple} divorce.} \\
& & {\color{Aquamarine} \{arg\}}'s marriage ended. \\
& & {\color{Aquamarine} \{arg\}} be {\color{RoyalPurple}hired.} \\
& & {\color{Aquamarine} \{arg\}} {\color{RoyalPurple} start a job.} \\
& & {\color{Aquamarine} \{arg\}} be {\color{RoyalPurple}fired.} \\
& & {\color{Aquamarine} \{arg\}} {\color{RoyalPurple} end a job.} \\
& & {\color{Aquamarine} \{arg\}} be {\color{RoyalPurple}nominated.} \\
& & {\color{Aquamarine} \{arg\}} be {\color{RoyalPurple}elected.} \\
& & {\color{Aquamarine} \{arg\}} be {\color{RoyalPurple}arrested.} \\
& & {\color{Aquamarine} \{arg\}} be {\color{RoyalPurple}jailed.} \\
& & {\color{Aquamarine} \{arg\}} be {\color{RoyalPurple}imprisoned.} \\
& & {\color{Aquamarine} \{arg\}} be {\color{RoyalPurple}released.} \\
& & {\color{Aquamarine} \{arg\}} be {\color{RoyalPurple}paroled.} \\
& & {\color{Aquamarine} \{arg\}} be {\color{RoyalPurple}executed.} \\
& & {\color{Aquamarine} \{arg\}} be {\color{RoyalPurple}extradited.} \\ \midrule

Place & {\color{RoyalPurple} \{trg\}} occurred in {\color{Aquamarine} \{arg\}}. & {\color{Aquamarine} \{arg\}} be a place. \\
&  & {\color{Aquamarine} \{arg\}} be a location. \\
& & {\color{Aquamarine} \{arg\}} be a placement. \\ \midrule

Plaintiff & {\color{Aquamarine} \{arg\}} {\color{RoyalPurple}filed suit against} {\color{Orange} someone}. & {\color{Aquamarine} \{arg\}} {\color{RoyalPurple}bring a lawsuit against} {\color{Orange} someone}.  \\ 
 & & {\color{Aquamarine} \{arg\}} {\color{RoyalPurple}bring a lawsuit against} {\color{Orange} something}.\\ 
 & & {\color{Aquamarine} \{arg\}} {\color{RoyalPurple}sue} {\color{Orange} someone}.\\ 
 & & {\color{Aquamarine} \{arg\}} {\color{RoyalPurple}sue} {\color{Orange} something}.\\    \midrule

Prosecutor& {\color{Aquamarine} \{arg\}} {\color{RoyalPurple}indicted} {\color{Orange}the defendant}.& {\color{Aquamarine} \{arg\}} prosecute. \\
& {\color{Aquamarine} \{arg\}} {\color{RoyalPurple}charged} {\color{Orange}the defendant}. & {\color{Aquamarine} \{arg\}} take somebody to court for a crime.\\ \midrule

Recipient &{\color{Aquamarine} \{arg\}} {\color{RoyalPurple}received} money from {\color{Orange} someone}. & {\color{Aquamarine} \{arg\}} {\color{RoyalPurple}receive} {\color{Orange} something}.\\
& & {\color{Aquamarine} \{arg\}} {\color{RoyalPurple}get} {\color{Orange} something}.\\
& & {\color{Aquamarine} \{arg\}} {\color{RoyalPurple}get} {\color{Orange}money}.\\ \midrule

Seller &{\color{Aquamarine} \{arg\}} {\color{RoyalPurple}sold} {\color{Orange} something}. & {\color{Aquamarine} \{arg\}} {\color{RoyalPurple}sell} {\color{Orange} something}.\\ \midrule

Target &{\color{Aquamarine} \{arg\}} was {\color{RoyalPurple} \{trg\_subtype\}}.& {\color{Aquamarine} \{arg\}} be {\color{RoyalPurple}attacked}.\\ 
& & {\color{RoyalPurple} \{trg\_type\}}'s target be {\color{Aquamarine} \{arg\}}.\\ \midrule

Vehicle & {\color{Aquamarine} \{arg\}} was used as a vehicle.& {\color{Aquamarine} \{arg\}} be a transport.\\
&   & {\color{Aquamarine} \{arg\}} be a vehicle.\\
&   & {\color{Aquamarine} \{arg\}} serve to move.\\
&   & {\color{Aquamarine} \{arg\}} serve to change location.\\
&   & {\color{Aquamarine} \{arg\}} serves as a means of transportation.\\ \midrule

Victim &  {\color{Aquamarine} \{arg\}} was {\color{RoyalPurple} \{trg\}}. & {\color{Aquamarine} \{arg\}} be victim.\\
&   & {\color{Aquamarine} \{arg\}} be {\color{RoyalPurple}injured}.\\
&   & {\color{Aquamarine} \{arg\}} be {\color{RoyalPurple}killed}.\\
&   & {\color{Aquamarine} \{arg\}} be {\color{RoyalPurple}harmed}.\\
&   & {\color{Aquamarine} \{arg\}} have a dead.\\
&   & {\color{Aquamarine} \{arg\}} have a tragedy.\\ \midrule

   \caption*{The templates written by both developers for ACE.}
    \label{tab:examplesTemplates}
\end{longtable*}

\clearpage
\twocolumn

\end{document}